\documentclass[letterpaper, 10 pt, conference]{ieeeconf}
\IEEEoverridecommandlockouts             
\usepackage{cite}
\usepackage{graphics} 
\usepackage{epsfig} 
\usepackage{times}
\usepackage{amsmath}
\usepackage{amssymb}
\usepackage{mathtools}
\usepackage[inkscapeformat=png]{svg}
\usepackage{algpseudocode}
\usepackage{algorithmicx}
\usepackage[linesnumbered, ruled, vlined]{algorithm2e}
\algnewcommand{\AND}{\algorithmicand}
\usepackage{booktabs}
\usepackage{multirow}
\usepackage{tabularx}
\usepackage{url}
\usepackage{balance}
\usepackage[textsize=footnotesize]{todonotes}

\title{\LARGE \bf FlipToSee: A Probabilistic Stable Placement Prior for Active Visual Exploration via Regrasping}
\author{Chang Shu, Sushil Samuel Dinesh, and Shinkyu Park
\thanks{The work was supported by funding from King Abdullah University of Science and Technology (KAUST).}
\thanks{The authors are with Electrical and Computer Engineering, King Abdullah University of Science and Technology (KAUST), Thuwal, 23955-6900, Saudi Arabia. Emails:
        {\tt\small \{chang.shu, sushilsamuel.dinesh, shinkyu.park\}@kaust.edu.sa }}%
}

\begin{document}
\maketitle
\thispagestyle{empty}
\pagestyle{empty}
\begin{abstract}
Active visual exploration of tabletop objects often requires reorienting an unknown resting object onto a different stable support face to expose occluded surfaces. To identify such placements without exhaustive physical search, we learn a probabilistic placement prior from a single-view point cloud. Stable placement prediction is inherently multimodal, and conventional 6-DoF regression introduces further ambiguity by modeling translation and in-plane yaw. We therefore propose FlipToSee, a probabilistic framework that removes this representational ambiguity by parameterizing placements as unit support normals on $S^2$ while modeling their multimodal conditional distribution via a von Mises--Fisher mixture density network. To decouple mode diversity from physical robustness, FlipToSee deterministically extracts a compact candidate set from the mixture components and applies robustness-aware reranking using an auxiliary head trained with candidate-aligned supervision. In simulation, FlipToSee achieves $98.4\%$ first-proposal success on in-distribution objects, $95.3\%$ on out-of-distribution shapes, and $90.0\%$ under zero-shot transfer to household YCB objects. We further demonstrate the learned placement prior on a physical robot by integrating it with grasp and motion planning for exploratory regrasping.\footnote{Code for implementation and training, along with the dataset used, will be released after the review process to ensure reproducibility.}

\end{abstract}


\section{Introduction} \label{sec:introduction}
\begin{figure}[htbp]
    \centering
    \includegraphics[width=\linewidth]{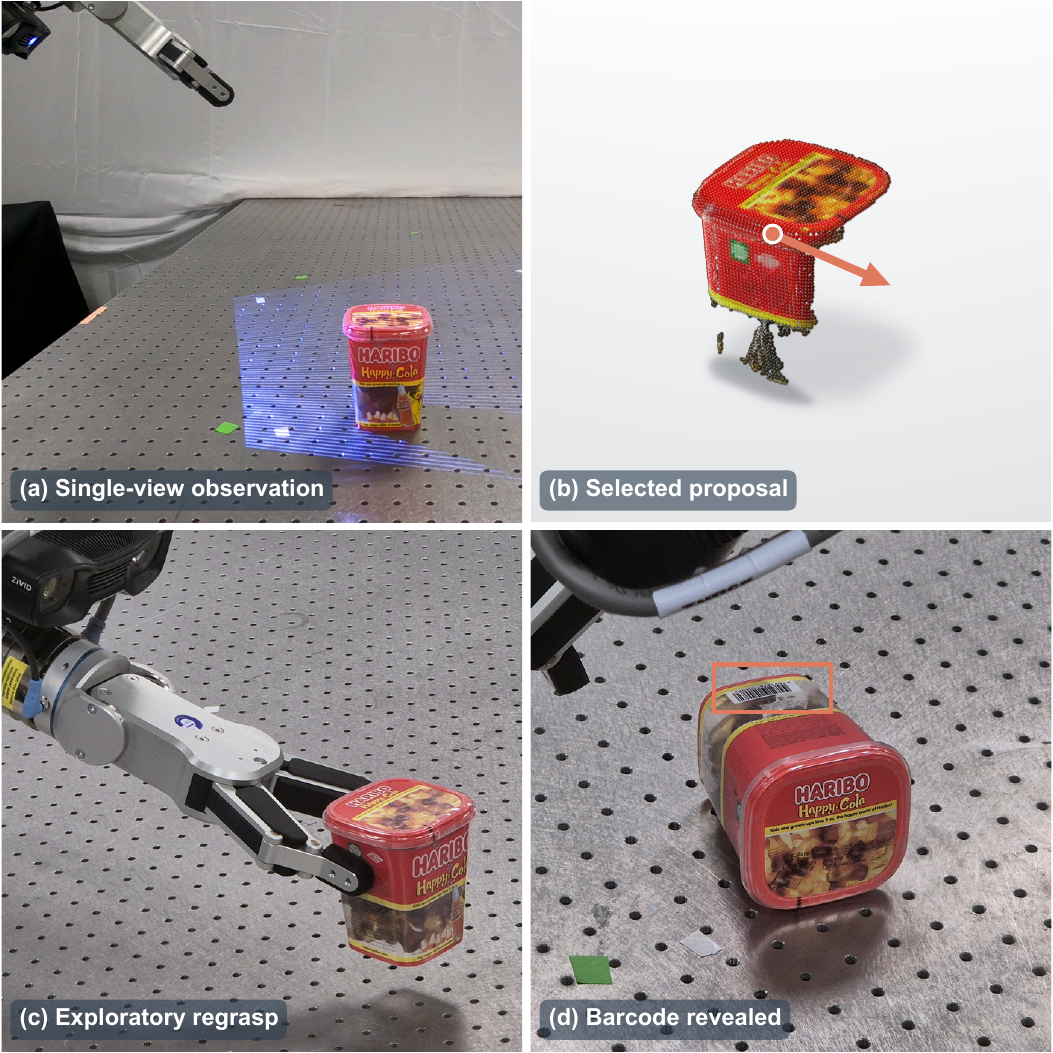}
    \caption{Executing a selected FlipToSee proposal during exploratory regrasping reveals a barcode occluded in the initial observation.}
    \label{fig:regrasp}
\end{figure}

Active visual exploration requires a robot to change what its camera can observe, either by selecting new camera viewpoints~\cite{9670705} or by reorienting an object~\cite{5980429,10757428,10801767}. In tabletop settings, self-occlusion and table contact obscure surfaces in the current resting pose. This motivates exploratory regrasping, where a robot observing an unknown object from a single viewpoint flips it to expose occluded surfaces. Different stable placements can expose complementary surfaces, and a stable pose is not always reachable via a feasible regrasp. To support both exploration and execution, we learn a multimodal placement prior from the partial observation, providing diverse placement proposals for downstream grasp and motion planners. As illustrated in Fig.~\ref{fig:regrasp}, flipping a snack box into one of these placements reveals features unavailable in the initial observation---such as a barcode.

Learning to predict alternative stable placements from a single-view observation presents three key challenges: (1)~\emph{Representation Redundancy}: Conventional placement methods often formulate the task as full 6-DoF pose prediction~\cite{cheng2021learning,nadeau2025generating}. In tabletop manipulation, however, translation and in-plane yaw are governed by graspability and kinematic constraints rather than placement stability; modeling them introduces unnecessary prediction ambiguity and scatters probability density across task-irrelevant degrees of freedom. (2)~\emph{Placement Multimodality}: The placement task is multimodal: a single observation admits multiple valid stable placements. Deterministic models collapse these modes by averaging them, often yielding predictions that correspond to no stable support face. While latent-variable generative models produce diverse hypotheses, they lack an explicit probability density; without a principled density ranking, identifying a valid placement among these hypotheses may require exhaustive physical search. (3)~\emph{Density--Robustness Misalignment}: An explicit density distinguishes candidates by placement likelihood but does not encode their physical robustness to angular perturbations; ranking candidates solely by density can therefore place a perturbation-sensitive candidate ahead of a more robust alternative.

To overcome these limitations, we propose \emph{FlipToSee}, a probabilistic framework that models the conditional distribution of stable placements directly on the $S^2$ support normal manifold---the continuous surface of the unit sphere. By employing a Mixture Density Network (MDN)~\cite{MDN} with von Mises--Fisher (vMF) distributions~\cite{mardia1999directional}, FlipToSee respects the directional constraint that support normals lie on $S^2$ and avoids the redundancies of full 6-DoF regression. To decouple mode diversity from physical robustness, our framework operates in two stages. First, we extract the component means with the highest density scores to form a compact candidate set. Second, a robustness head trained with candidate-aligned supervision estimates each candidate's perturbation robustness, enabling robustness-aware reranking based on the density and robustness scores.

Our contributions are as follows:
\begin{itemize}
    \item We formulate the prediction of alternative stable placements as a continuous density estimation problem on the $S^2$ support normal manifold, simultaneously addressing placement multimodality and directional constraints.
    \item We develop a placement prior based on a vMF-MDN to model the multimodal distribution of support normals as an explicit continuous density, coupled with deterministic mode extraction to derive a compact candidate set from component mean directions.
    \item We present a reranking strategy that combines density scores with robustness scores learned via candidate-aligned supervision, prioritizing robust placements without sacrificing mode coverage.
\end{itemize}

The remainder of this paper reviews related work (Section~\ref{sec:related-work}), formulates the stable placement problem (Section~\ref{sec:formulation}), and details the FlipToSee framework (Section~\ref{sec:method}). Experiments and a robotic demonstration are presented in Section~\ref{sec:experiments}, followed by the conclusion in Section~\ref{sec:conclusion}.

\begin{figure*}[htbp]
    \centering
    \includegraphics[width=\linewidth]{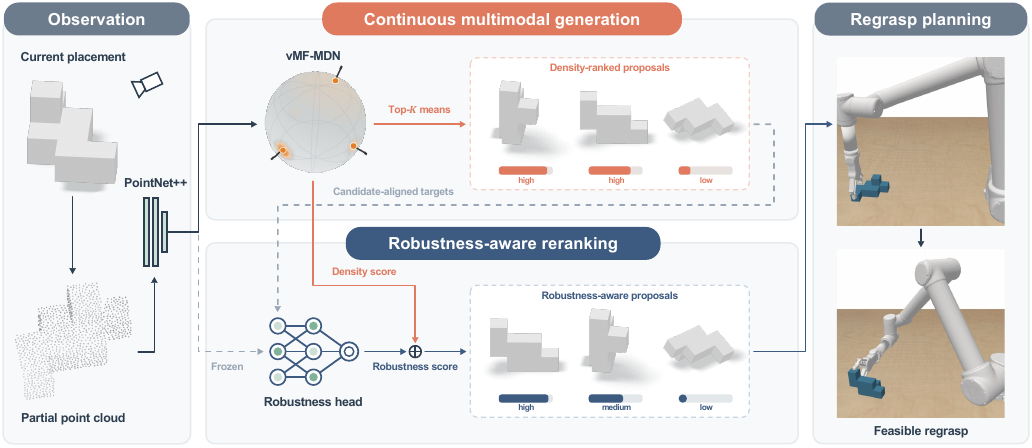}
    \caption{The FlipToSee framework. A point cloud encoder maps the partial point cloud to a global shape feature, which conditions an MDN that predicts a continuous vMF mixture over the $S^2$ support normal manifold. Deterministic mode extraction selects the $K$ component means with the highest density scores as a candidate set. Finally, robustness-aware reranking combines the density score with a robustness score from an auxiliary head trained with candidate-aligned supervision. The regrasp planner then evaluates the ranked proposals for graspability and kinematic feasibility.}
    \label{fig:method_overview}
\end{figure*}

\section{Related Work} \label{sec:related-work}

\subsection{Geometry-Based Placement Planning}
Geometry-based placement planners search for poses that satisfy support, collision, and manipulation constraints. Haustein et al.~\cite{8967732} jointly optimize stable placements and their reachability by a robot manipulator, and Mitash et al.~\cite{9131812} couple task-driven perception with constrained placement of unknown objects. More recently, Nadeau and Kelly~\cite{11027417} select robust contact points before solving for a compatible placement pose, reversing the conventional sample-and-evaluate pipeline. These methods provide explicit physical reasoning and accommodate general scene constraints, but their goal is to compute individual feasible poses via online geometric search. In contrast, we learn a compact prior over alternative support directions from a single-view observation and retain graspability and kinematic validation for a regrasp planner.

\subsection{Learning Stable Placements from Point Clouds}
Learning-based methods replace hand-designed placement search with predictors that generalize across object geometry. Jiang et al.~\cite{doi:10.1177/0278364912438781} introduced a placing score based on hand-crafted geometric and contextual features, and subsequent methods learn placement-relevant features directly from 3D observations using point cloud encoders such as PointNet++~\cite{NIPS2017_d8bf84be} and DGCNN~\cite{10.1145/3326362}. The closest prior work in terms of observation setting is UOP-Net~\cite{uopnet}, which segments the most stable support plane of an unseen object from a single-view point cloud, without assuming the object begins in a known stable placement. Our setting differs in both starting condition and objective: we assume the object is already resting stably, and rather than selecting a single preferred support plane for initial placement, we predict multiple alternative support normals---since exploratory regrasping must expose surfaces currently occluded by the object's resting placement.

\subsection{Multimodal Placement Prediction and Regrasping}
Stable placement prediction is multimodal, motivating models that generate sets or distributions of poses. Cheng et al.~\cite{cheng2021learning} condition a latent-code generator on partial object and environment point clouds to produce diverse 6-DoF placements for a regrasp graph. Xu et al. use generative networks to predict diverse stable orientations on a planar support~\cite{10313307} and feasible orientations afforded by extrinsic supports~\cite{10258116} for object reorientation. Paxton et al.~\cite{paxton2021predicting} learn an MDN prior over 3D translation and planar rotation for semantic placement; this parameterization preserves the current support normal and therefore cannot select a new support face. Nadeau et al.~\cite{nadeau2025generating} learn a scene-conditioned diffusion model over full 6-DoF poses, but address where and how to place an object rather than how to reorient an already stable object for exploration. We instead model the latter decision as a vMF mixture over support normals on $S^2$, producing diverse, ranked placement proposals for regrasping without modeling translation or yaw.

\section{Problem Formulation} \label{sec:formulation}
We consider a rigid object resting on a planar table. Let $\mathbf{X} \in \mathbb{R}^{N \times 3}$ denote a single-view point cloud of the object, where $N$ is the number of points. We normalize $\mathbf{X}$ to zero mean and unit scale so that the observation is independent of the object's absolute position and metric size. Given only this normalized partial observation, and without prior knowledge of the object's stable orientations, our task is to predict support normals that, when the object is reoriented accordingly, place it on an alternative stable support face---thereby exposing surfaces occluded in the current placement.

We parameterize each stable placement by its support normal, which must align with gravity at equilibrium. Given an observation $\mathbf{X}_i$ at the current placement $i$, let $\mathbf{R}_i, \mathbf{R}_j \in SO(3)$ denote the object orientations at stable placements $i$ and $j$, respectively, represented as rotations from the object frame to the observation frame of $\mathbf{X}_i$, and let $\mathbf{g}=[0,0,-1]^\top$ be the unit gravity direction. The support normal for placement $j$, expressed in the observation frame, is
\begin{equation*}
    \mathbf{n}_{ij}=\mathbf{R}_i\mathbf{R}_j^\top\mathbf{g},
    \qquad \mathbf{n}_{ij}\in S^2.
\end{equation*}
Because planar translation and yaw do not alter the active support face, this $S^2$ directional representation isolates the minimal degrees of freedom required to specify a stable support face. Accordingly, the set of valid alternative support normals for observation $\mathbf{X}_i$ is defined as
\begin{equation*}
    \mathcal{N}_i=\left\{\mathbf{n}_{ij}\mid j\neq i\right\},
\end{equation*}
where $j\neq i$ excludes the object's current placement.

Given training pairs $(\mathbf{X}_i,\mathcal{N}_i)$, our objective is to learn a conditional distribution $p_\theta(\mathbf{n}\mid\mathbf{X}_i)$ on $S^2$ that assigns high density to all valid alternative directions. During inference, given a verification budget $K$, a prediction method produces a ranked list of proposals $\widehat{\mathcal{N}}_i=(\hat{\mathbf{n}}_{i,1},\ldots,\hat{\mathbf{n}}_{i,K})$. For simulation evaluation, these proposals are verified in rank order until one succeeds.

To verify a proposal $\hat{\mathbf{n}}_{i,k}$, we rotate the object to align $\hat{\mathbf{n}}_{i,k}$ with $\mathbf{g}$ and simulate its physical placement (see Section~\ref{sec:datasets} for simulation details). Let $\mathbf{R}^{\mathrm{set}}_{i,k}$ denote the settled orientation; the corresponding support normal in the observation frame of $\mathbf{X}_i$ is
\begin{equation*}
    \mathbf{n}^{\mathrm{set}}_{i,k}
    =\mathbf{R}_i(\mathbf{R}^{\mathrm{set}}_{i,k})^\top\mathbf{g}.
\end{equation*}
A prediction succeeds if $(i)$ the object reaches a static equilibrium, $(ii)$ $\mathbf{n}^{\mathrm{set}}_{i,k}$ remains within an angular tolerance $\epsilon$ of $\hat{\mathbf{n}}_{i,k}$, and $(iii)$ $\mathbf{n}^{\mathrm{set}}_{i,k}$ deviates from $\mathbf{g}$ by more than $\epsilon$. The goal is to maximize the probability of finding a valid alternative placement within the verification budget $K$, while minimizing the required number of checks. Note that this formulation focuses exclusively on placement stability, leaving graspability and kinematic constraints to a downstream regrasp planner; visibility-aware placement selection falls outside the scope of this work.

\section{Method} \label{sec:method}
The proposed FlipToSee framework infers a ranked list of alternative stable placement proposals from a single-view observation. To decouple mode diversity from physical robustness, our architecture separates continuous multimodal generation from robustness-aware reranking. As illustrated in Fig.~\ref{fig:method_overview}, our placement prior---comprising a point cloud encoder and an MDN---maps the partial point cloud $\mathbf{X}_i$ to a global shape feature and subsequently to a continuous vMF mixture over the $S^2$ support normal manifold. Rather than relying on stochastic sampling, deterministic mode extraction selects the $K$ component means with the highest density scores as a compact candidate set. Robustness-aware reranking then combines the density score with a robustness score from an auxiliary head trained with candidate-aligned supervision. The resulting ranked proposals are passed to the regrasp planner, which plans a feasible regrasp subject to graspability and kinematic constraints.

\subsection{Directional Modeling of Multimodal Placements}
We encode the partial point cloud $\mathbf{X}_i$ with a PointNet++ backbone~\cite{NIPS2017_d8bf84be}, extracting a global shape feature $\mathbf{h}_i\in\mathbb{R}^{512}$. A multi-layer perceptron (MLP) mixture head $f_\theta: \mathbb{R}^{512} \to \mathbb{R}^{5M}$ then maps $\mathbf{h}_i$ to the concatenated parameters of $M$ vMF components, translating the geometric encoding into a continuous probability distribution over stable placements.

Formally, we model the conditional density of a unit support normal $\mathbf{n}\in S^2$ as a vMF mixture:
\begin{equation*}
    p_\theta(\mathbf{n}\mid\mathbf{X}_i)
    =\sum_{m=1}^{M}\pi_{i,m}
    C_3(\kappa_{i,m})
    \exp\!\left(
        \kappa_{i,m}\boldsymbol{\mu}_{i,m}^{\top}\mathbf{n}
    \right),
\end{equation*}
where the scalar weight $\pi_{i,m}\in(0,1)$, the 3D mean direction $\boldsymbol{\mu}_{i,m}\in S^2$, and the scalar concentration $\kappa_{i,m}>0$ denote the parameters of component $m$, with $\sum_{m=1}^M \pi_{i,m}=1$. The normalizing constant on the 2-sphere is given by $C_3(\kappa)=\kappa/(4\pi\sinh\kappa)$~\cite{mardia1999directional}. To obtain valid mixture parameters, we apply a softmax to the weight logits, $L_2$-normalize the mean-direction outputs, and map the concentration outputs to a finite positive range using a scaled sigmoid. By modeling support normals on $S^2$ rather than in $\mathbb{R}^3$, this vMF formulation avoids scattering probability density away from the unit sphere. Each $\boldsymbol{\mu}_{i,m}$ defines a candidate support normal, and $\kappa_{i,m}$ determines the angular sharpness around that mode.

\subsection{Mode Extraction and Robustness-Aware Reranking}
Given a limited verification budget, we cannot afford redundant candidates or an arbitrary verification order; instead, we must extract a compact yet diverse candidate set, ranked so that the most promising proposals are verified first. Standard stochastic sampling from a continuous mixture can waste this budget by drawing redundant normals from a few dominant peaks. To bypass sampling variance, we use deterministic mode extraction to expose the hypotheses represented by the mixture components.

Although exact mode-seeking in a vMF mixture is analytically intractable, each mean direction $\boldsymbol{\mu}_{i,m}$ is the unique mode of its component and a natural representative of its directional hypothesis. For each component mean, we compute the density score $s_{\mathrm{density}}(\boldsymbol{\mu}_{i,m}) = \log\pi_{i,m} + \log C_3(\kappa_{i,m}) + \kappa_{i,m}$, which incorporates both mixture weight and angular concentration. We then select the $K$ ($K \le M$) component means with the highest scores to form a candidate set $\mathcal{C}_i=\{\mathbf{c}_{i,1},\ldots,\mathbf{c}_{i,K}\}$. This extraction yields one candidate per prominent directional hypothesis, with the resulting candidates ranked by their density scores.

Density ranking reflects the placement likelihood learned by the placement prior, but does not account for robustness to angular perturbations: a candidate associated with a narrow support face may fail under slight execution errors, yet the density score does not explicitly distinguish it from a candidate on a more robust support face. Moreover, a predicted candidate can exhibit angular deviations from its nearest stable mode. To incorporate physical robustness without sacrificing mode coverage, we introduce a lightweight robustness head $g_{\phi}: \mathbb{R}^{515} \to \mathbb{R}$ for reranking the candidate set $\mathcal{C}_i$. Implemented as an MLP, this auxiliary network processes the global shape feature $\mathbf{h}_i$, extracted from the placement prior, alongside a candidate normal $\mathbf{c} \in \mathcal{C}_i$ to predict a robustness estimate:
\begin{equation*}
    q_{\phi}(\mathbf{c}\mid\mathbf{X}_i)
    = \sigma\!\left(g_{\phi}([\mathbf{h}_i;\mathbf{c}])\right).
\end{equation*}

At inference, we define the robustness score as $s_{\mathrm{robust}}(\mathbf{c}\mid\mathbf{X}_i)=\log q_{\phi}(\mathbf{c}\mid\mathbf{X}_i)$. The sum of the density and robustness scores gives the joint score $s_{\mathrm{joint}}(\mathbf{c}\mid\mathbf{X}_i)=s_{\mathrm{density}}(\mathbf{c})+s_{\mathrm{robust}}(\mathbf{c}\mid\mathbf{X}_i)$. Ranking candidates in descending order by this score establishes \emph{joint ranking}, which prioritizes candidates that are both likely and robust to angular perturbations. The reranked candidates form the proposal list $\widehat{\mathcal{N}}_i=(\hat{\mathbf{n}}_{i,1},\ldots,\hat{\mathbf{n}}_{i,K})$, with overall mode coverage unchanged.

\subsection{Generative and Candidate-Aligned Training}
We train the FlipToSee framework in two stages: first optimizing the placement prior on mode-balanced directional targets, and subsequently training the robustness head on candidate-aligned robustness targets.

When training generative placement models, directly utilizing raw, unbalanced simulation outcomes---where modes with larger stability regions yield more positive samples---would weight dominant support faces more heavily than less frequent modes, even though the latter could provide useful alternatives for both exploration and execution. To prevent such sampling bias from degrading mode diversity, we train the placement prior on a mode-balanced target set $\mathcal{N}_i$. As detailed in Section~\ref{sec:datasets}, $\mathcal{N}_i$ is constructed by clustering continuous stable poses into discrete modes and retaining one representative normal per mode. We optimize the prior's parameters $\theta$ by minimizing the negative log-likelihood:
\begin{equation*}
    \mathcal{L}_{\mathrm{vMF}}
    =-\frac{1}{\lvert\mathcal{N}_i\rvert}
    \sum_{\mathbf{n}\in\mathcal{N}_i}
    \log p_\theta(\mathbf{n}\mid\mathbf{X}_i).
\end{equation*}

In the second stage, we freeze the placement prior and train the robustness head on the same partial point clouds. We first quantify the physical robustness of each ground-truth mode $\mathbf{n}\in\mathcal{N}_i$ as the fraction of perturbed placements that settle back into $\mathbf{n}$. However, these robustness values are defined only for the ground-truth modes, whereas the auxiliary head estimates each candidate's robustness in $\mathcal{C}_i$ at inference. Because such candidates can deviate from their nearest stable mode, training only on the ground-truth modes would leave them without direct supervision.

We therefore construct candidate-aligned targets $y_i(\mathbf{c})$ by mapping each candidate $\mathbf{c}\in\mathcal{C}_i$ to its nearest ground-truth normal $\mathbf{n}^{*}(\mathbf{c})$. If the resulting angular deviation falls within the tolerance $\epsilon$, the candidate inherits its nearest mode's robustness fraction as its target $y_i(\mathbf{c})$; otherwise, it is assigned $y_i(\mathbf{c}) = 0$. This target construction provides hard negative supervision for candidates outside the angular tolerance while preserving soft robustness supervision for candidates matched to stable modes. Reusing precomputed mode robustness provides an efficient offline proxy, avoiding repeated perturbation simulations for every candidate during training. We optimize the robustness parameters $\phi$ by minimizing the binary cross-entropy (BCE) loss:
\begin{equation*}
    \mathcal{L}_{\mathrm{robust}}
    = \frac{1}{\lvert\mathcal{C}_i\rvert}
    \sum_{\mathbf{c}\in\mathcal{C}_i}
    \operatorname{BCE}\!\left(q_{\phi}(\mathbf{c}\mid\mathbf{X}_i),\, y_i(\mathbf{c})\right).
\end{equation*}

\section{Experiments} \label{sec:experiments}

\subsection{Experimental Setup}
\subsubsection{Datasets and Simulation Environment} \label{sec:datasets}
\begin{figure}[t]
    \centering
    \includegraphics[width=0.9\linewidth]{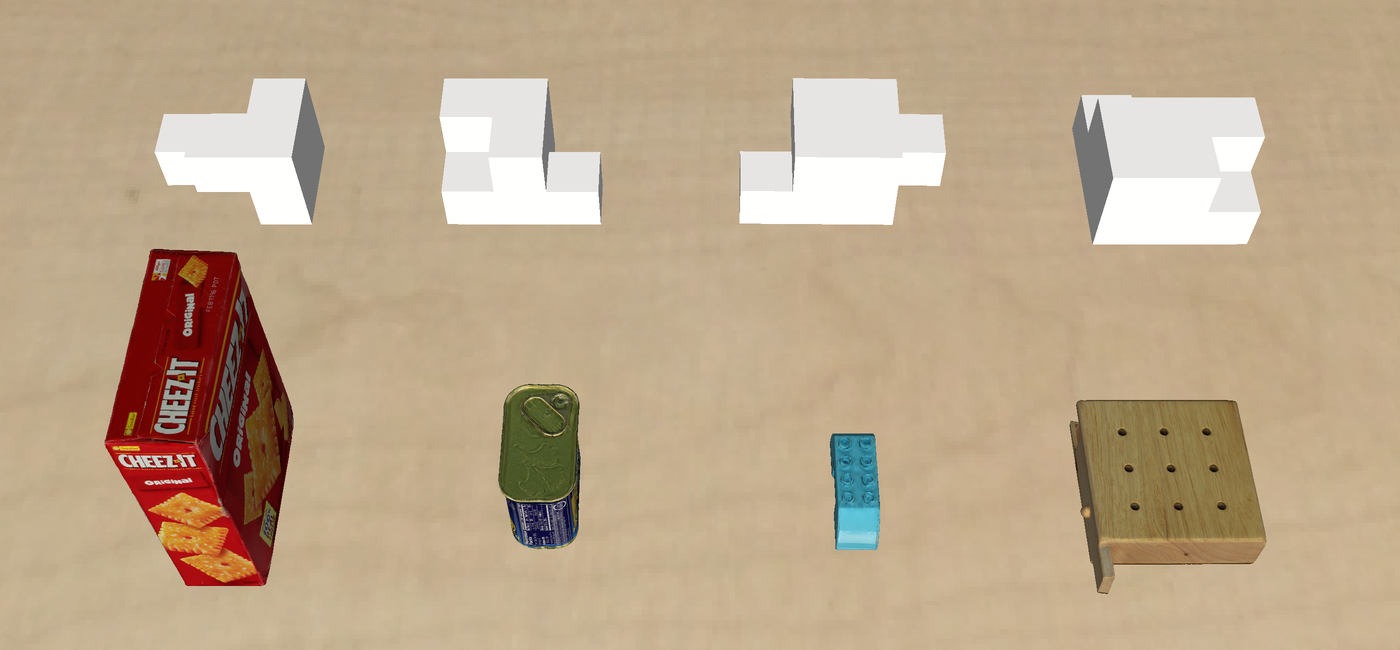}
    \caption{Representative objects from our datasets. \textbf{Top:} Block objects containing 7, 8, 9, and 10 voxels. \textbf{Bottom:} YCB objects used for zero-shot transfer evaluation.}
    \label{fig:datasets}
\end{figure}

We evaluate on synthetic block sets and the YCB object set~\cite{7251504} (see Fig.~\ref{fig:datasets}). Each block object is a connected polycube generated by starting from one unit cube and iteratively attaching cubes to exposed faces. The block sets span four voxel counts $\{7, 8, 9, 10\}$. Within each count, we canonicalize shapes up to translation and the 24 orientation-preserving rotations\footnote{The $3\times3$ signed permutation matrices with determinant $+1$.}, retaining a single representative to prevent duplicates across data splits. For each voxel count in $\{7, 8, 9\}$, we generate $400$ objects, split into $70\%$ training, $15\%$ validation, and $15\%$ test data. Models are trained on the combined training data and evaluated on the held-out test data, forming our \textbf{In-Distribution (ID)} test set. To test extrapolation to larger polycubes, the $10$-voxel blocks ($60$ objects) are reserved as our \textbf{Out-of-Distribution (OOD)} test set. Finally, the YCB object set is used to evaluate \textbf{Zero-Shot} transfer from synthetic blocks to household geometries.

We generate ground-truth stable placements in PyBullet~\cite{pybullet} by dropping each object from $1{,}000$ random initial orientations with a $1$~cm clearance above the tabletop. A pose is accepted as stable if, within $10$~s, the object maintains table contact for $1$~s while its linear and angular velocities remain below $10^{-3}$~m/s and $10^{-2}$~rad/s, respectively. After clustering poses whose support normals differ by less than $\epsilon=10^\circ$ into stable modes, we evaluate each mode's robustness to $16$ angular perturbations, formed by combining four tilt angles $\{2.5^\circ,5^\circ,7.5^\circ,10^\circ\}$ with four evenly spaced tilt directions in the horizontal plane. A perturbation succeeds if the object reaches static equilibrium and its settled support normal differs by less than $\epsilon$ from both the perturbed target normal and the original mode normal. To exclude objects dominated by rolling behavior, we retain only YCB objects with at least two major stable modes---filtered by drop frequency and support area---that collectively account for at least $80\%$ of stable drops, yielding $24$ objects for zero-shot evaluation.

At each stable orientation, we render partial point clouds from $32$ camera poses sampled along a spherical Fibonacci lattice~\cite{gonzalez2010measurement} on an upper hemisphere of radius $0.5$~m, downsampling each view to $N=1{,}024$ points. During training, we sample a random view and apply Gaussian point jitter alongside a continuous gravity-axis rotation that identically transforms target normals. To ensure consistent evaluation across objects, we use four fixed, spatially distributed views.

\subsubsection{Evaluation Metrics}
Given an observation $\mathbf{X}_i$, each method produces a ranked list of $K$ proposals $\widehat{\mathcal{N}}_i$. For each $k\in\{1,\ldots,K\}$, we report the following metrics:
\begin{itemize}
    \item \textbf{Success@$k$} measures placement reliability as the percentage of test trials where at least one of the top-$k$ proposals yields a successful placement (per Section~\ref{sec:formulation}).
    \item \textbf{Checks@$k$} quantifies search efficiency as the average number of checks required to find a successful placement, capped at $k$ if all top-$k$ proposals fail.
    \item \textbf{Recall@$k$} evaluates mode coverage as the percentage of ground-truth stable modes in $\mathcal{N}_i$ recovered by the top-$k$ proposals, computed via maximum bipartite matching under angular tolerance $\epsilon$.
\end{itemize}

To prevent objects with many stable placements from dominating the evaluation, all metrics are averaged hierarchically: first across the four views per placement, then across placements per object, and finally across objects.

\subsubsection{Baseline Methods}
We benchmark vMF-MDN against three geometry-based heuristics and three learning-based baselines. The geometry-based methods operate directly on the observed point clouds:
\begin{itemize}
    \item \textbf{Convex hull stability analysis (CHSA)}~\cite{8967732} computes stable poses of the observed convex hull and ranks support normals by landing probability.
    \item \textbf{Bounding box fitting (BBF)}~\cite{9131812} ranks oriented-bounding-box face normals by face area.
    \item \textbf{RANSAC plane fitting (RPF)}~\cite{10.1145/358669.358692} iteratively extracts dominant planes and ranks outward-facing normals by inlier count.
\end{itemize}
The learning-based baselines use the same encoder architecture, input point clouds, and data splits as our method:
\begin{itemize}
    \item \textbf{MLP regression (MLP)} predicts a single unit normal by minimizing the mean cosine loss across all valid target normals.
    \item \textbf{Latent set generator (LSG)}, adapted from Cheng et al.~\cite{cheng2021learning}, maps latent codes and point cloud features to a set of normal hypotheses under a symmetric cosine-Chamfer loss.
    \item \textbf{Gaussian mixture density network (GMM-MDN)}, adapted from Paxton et al.~\cite{paxton2021predicting}, models support normals via a Gaussian mixture in $\mathbb{R}^3$, with component means normalized onto $S^2$ at inference.
\end{itemize}

\begin{table*}[t]
\centering
\caption{Stable placement prediction on Block ID and Block OOD. Both mixture models use density ranking.}
\label{tab:main_results}
\scriptsize
\setlength{\tabcolsep}{3pt}
\begin{tabularx}{\textwidth}{@{}l*{8}{>{\centering\arraybackslash}X}@{}}
\toprule
\multirow{2}{*}{Method} & \multicolumn{4}{c}{Block ID: 7--9 voxels} & \multicolumn{4}{c}{Block OOD: 10 voxels} \\
\cmidrule(lr){2-5}\cmidrule(lr){6-9}
& Success@1 & Checks@5 & Recall@1 & Recall@5 & Success@1 & Checks@5 & Recall@1 & Recall@5 \\
\midrule
CHSA~\cite{8967732}
& $58.91 \pm 0.00$ & $1.44 \pm 0.00$ & $8.16 \pm 0.00$ & $46.79 \pm 0.00$
& $57.96 \pm 0.00$ & $1.49 \pm 0.00$ & $7.34 \pm 0.00$ & $42.07 \pm 0.00$ \\
\addlinespace
BBF~\cite{9131812}
& $22.70 \pm 0.00$ & $3.72 \pm 0.00$ & $3.47 \pm 0.00$ & $10.22 \pm 0.00$
& $18.20 \pm 0.00$ & $3.82 \pm 0.00$ & $2.57 \pm 0.00$ & $7.84 \pm 0.00$ \\
\addlinespace
RPF~\cite{10.1145/358669.358692}
& $77.34 \pm 0.02$ & $1.35 \pm 0.00$ & $10.41 \pm 0.00$ & $32.35 \pm 0.00$
& $63.27 \pm 0.02$ & $1.52 \pm 0.00$ & $7.79 \pm 0.00$ & $27.17 \pm 0.02$ \\
\midrule
MLP
& $9.03 \pm 0.49$ & $4.64 \pm 0.02$ & $1.53 \pm 0.07$ & $1.53 \pm 0.07$
& $9.42 \pm 0.93$ & $4.62 \pm 0.04$ & $1.51 \pm 0.17$ & $1.51 \pm 0.17$ \\
\addlinespace
LSG~\cite{cheng2021learning}
& $59.43 \pm 1.54$ & $1.73 \pm 0.04$ & $8.65 \pm 0.19$ & $32.61 \pm 0.72$
& $50.75 \pm 0.96$ & $2.02 \pm 0.05$ & $6.95 \pm 0.15$ & $26.84 \pm 0.48$ \\
\addlinespace
GMM-MDN~\cite{paxton2021predicting}
& $70.21 \pm 2.22$ & $1.49 \pm 0.02$ & $8.83 \pm 0.43$ & $38.32 \pm 1.22$
& $61.73 \pm 2.54$ & $1.64 \pm 0.04$ & $7.03 \pm 0.27$ & $31.12 \pm 0.69$ \\
\addlinespace
\textbf{vMF-MDN (Ours)}
& $\mathbf{96.93 \pm 0.72}$ & $\mathbf{1.05 \pm 0.01}$ & $\mathbf{12.70 \pm 0.06}$ & $\mathbf{52.18 \pm 0.53}$
& $\mathbf{93.25 \pm 0.67}$ & $\mathbf{1.10 \pm 0.01}$ & $\mathbf{11.25 \pm 0.11}$ & $\mathbf{43.73 \pm 0.81}$ \\
\bottomrule
\end{tabularx}
\end{table*}

\subsubsection{Implementation Details} \label{sec:impl_details}
Both mixture models contain $M=14$ components, matching the 95th percentile of alternative stable mode counts in the Block ID training set. Each GMM component has a learned diagonal covariance. For numerical stability, we cap the vMF concentration at $150$ and lower bound each GMM standard deviation at $0.01$. For GMM-MDN, the density score combines the mixture weight and Gaussian density evaluated at the component mean. The same angular tolerance $\epsilon=10^\circ$ applies to physical verification, mode clustering, and candidate alignment, and the verification budget is $K=5$.

We train all networks in PyTorch~\cite{pytorch} on a single NVIDIA RTX 2080 Ti for up to $2{,}000$ epochs using Adam, with a batch size of $16$ and an initial learning rate of $2\times10^{-4}$. The learning rate is halved when the validation loss plateaus. For the placement prior, the scheduler patience and early stopping patience are set to $50$ and $150$ epochs, respectively. We shorten these values to $20$ and $50$ epochs for the lightweight robustness head, which converges faster. Finally, we adopt joint ranking for the full FlipToSee framework, as it produces the highest Success@1 on the Block ID validation set. Unless otherwise specified, simulation results are reported as the mean $\pm$ standard deviation over three random seeds.

\subsection{Results}
We first evaluate the placement prior under density ranking, then examine the effects of ranking criteria and robustness supervision, and finally assess the full FlipToSee framework on zero-shot YCB objects.

On both Block ID and Block OOD, explicit density models outperform the deterministic MLP and latent-variable LSG baselines (Table~\ref{tab:main_results}). The single-output MLP cannot represent diverse placement hypotheses, whereas LSG generates multiple hypotheses without an explicit density for ranking. By explicitly parameterizing the multimodal distribution, GMM-MDN raises Success@1 above $61\%$ on both Block sets.
Beyond mixture modeling, learning a distribution over support normals natively on $S^2$ rather than in $\mathbb{R}^3$ provides a representational advantage. Our vMF-MDN attains over $93\%$ Success@1 on both Block sets and exceeds GMM-MDN by more than $12$ points in Recall@5. With density ranking, vMF-MDN reduces the average number of checks to approximately $1.1$, showing that its density scores reliably prioritize stable support faces.

\begin{table}[t]
\centering
\caption{Effects of ranking criteria and robustness supervision on Success@1. Density, Robustness, and Joint rank the same candidates by $s_{\mathrm{density}}$, $s_{\mathrm{robust}}$, and $s_{\mathrm{joint}}$, respectively.}
\label{tab:robustness_ablation}
\scriptsize
\setlength{\tabcolsep}{2.5pt}
\begin{tabular*}{\columnwidth}{@{\extracolsep{\fill}}lccc@{}}
\toprule
Model \& Variant & Block ID & Block OOD & YCB \\
\midrule
GMM-MDN (Density) & $70.21 \pm 2.22$ & $61.73 \pm 2.54$ & $85.13 \pm 6.29$ \\
\quad Robustness & & & \\
\qquad Mode-supervised & $75.15 \pm 1.66$ & $63.22 \pm 1.16$ & $83.80 \pm 9.55$ \\
\qquad Candidate-aligned & $78.42 \pm 2.97$ & $69.41 \pm 3.94$ & $82.64 \pm 9.36$ \\
\quad Joint & & & \\
\qquad Mode-supervised & $73.29 \pm 2.13$ & $63.34 \pm 2.08$ & $85.21 \pm 6.13$ \\
\qquad Candidate-aligned & $77.58 \pm 2.33$ & $67.25 \pm 3.72$ & $84.87 \pm 7.46$ \\
\midrule
vMF-MDN (Density) & $96.93 \pm 0.72$ & $93.25 \pm 0.67$ & $80.70 \pm 3.64$ \\
\quad Robustness & & & \\
\qquad Mode-supervised & $96.55 \pm 0.33$ & $90.77 \pm 0.49$ & $88.33 \pm 2.70$ \\
\qquad Candidate-aligned & $98.22 \pm 0.08$ & $94.04 \pm 0.46$ & $\mathbf{91.15 \pm 2.87}$ \\
\quad Joint & & & \\
\qquad Mode-supervised & $97.94 \pm 0.48$ & $94.23 \pm 0.32$ & $87.39 \pm 3.97$ \\
\qquad \textbf{Candidate-aligned} & $\mathbf{98.35 \pm 0.23}$ & $\mathbf{95.28 \pm 0.74}$ & $90.02 \pm 4.20$ \\
\bottomrule
\end{tabular*}
\end{table}

For vMF-MDN, joint ranking achieves the highest Success@1 on both Block sets under either supervision strategy, exceeding density ranking by about $1$--$2$ points (Table~\ref{tab:robustness_ablation}). These gains suggest that the density and robustness scores provide complementary signals for first-proposal selection on the Block sets. On YCB, however, robustness ranking exceeds joint ranking by about one point under either supervision strategy, possibly reflecting fewer stable modes among YCB objects and a correspondingly greater benefit from prioritizing robustness. Despite this gap, joint ranking remains about $7$--$9$ points above density ranking. Candidate-aligned supervision further improves vMF-MDN under both robustness and joint ranking. Across these ranking criteria, this alignment increases Success@1 in all six comparisons by $0.4$--$3.3$ points, indicating that candidate-aligned supervision is well suited to the vMF parameterization.

However, the results for GMM-MDN reveal a different interaction between density and robustness. Robustness-aware reranking improves Success@1 on both Block sets by approximately $1.5$--$8$ points under either supervision strategy, but offers no clear benefit on YCB. With candidate-aligned supervision on YCB, robustness ranking falls about $2.5$ points below density ranking; combining robustness with density in the joint score narrows this gap to $0.3$ points. This pattern suggests that the GMM robustness estimates transfer less consistently to household objects.

\begin{figure}[htbp]
    \centering
    \includegraphics[width=\columnwidth]{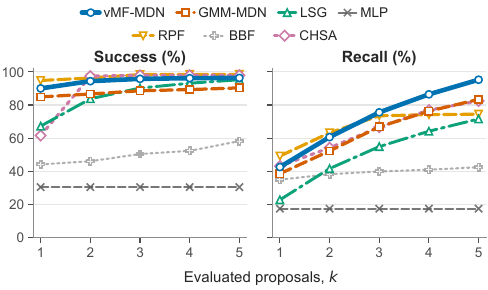}
    \caption{Zero-shot YCB performance as the number of evaluated proposals $k$ increases. For a controlled comparison, both mixture models use joint ranking with candidate-aligned supervision.}
    \label{fig:main_results}
\end{figure}

Fig.~\ref{fig:main_results} compares the full FlipToSee framework---featuring joint ranking with candidate-aligned supervision---against the baselines on zero-shot YCB. RPF obtains the highest first-attempt success, consistent with its preference for large planar surfaces. While favoring placement stability, this geometric bias restricts mode coverage, with Recall@$k$ plateauing near $75\%$. For active visual exploration, limited coverage reduces the placement options for exposing occluded surfaces. During execution, placing an object on its largest planar surface can also be impractical---it can force the gripper into configurations that collide with the tabletop during placement or subsequent regrasps.

Together, these requirements motivate a diverse set of alternative stable placements. CHSA attempts to enumerate all resting poses via the object's convex hull but is sensitive to single-view occlusion; moreover, ranking by landing probability rather than learned placement priors results in lower first-attempt success. In contrast, FlipToSee addresses the limitations of these geometry-based heuristics: RPF covers fewer modes and CHSA lacks robustness-aware reranking, whereas FlipToSee achieves over $95\%$ Recall@5 while maintaining around $90\%$ Success@1.

\subsection{Robotic Demonstration}

\begin{figure}[t]
    \centering
    \includegraphics[width=\linewidth]{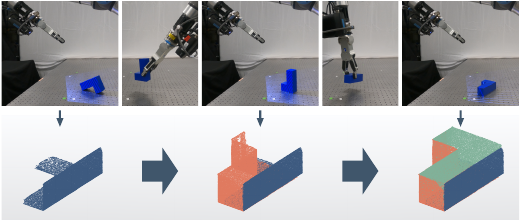}
    \caption{Active visual exploration of an L-shaped block on a physical robot. \textbf{Top:} Two successive regrasps. \textbf{Bottom:} Progressive accumulation of the three registered point clouds.}
    \label{fig:active_visual_exploration}
\end{figure}

We deploy the vMF-MDN placement prior in an active visual exploration task on a physical robot (Fig.~\ref{fig:active_visual_exploration}). Our platform consists of a Universal Robots UR10e manipulator equipped with an OnRobot RG6 two-finger gripper and a Zivid~2 structured-light 3D camera mounted on the manipulator. The robot reorients an L-shaped block to expose surfaces occluded by its current pose and table contact. For this demonstration, we use the placement prior without fine-tuning and apply density ranking to form the candidate set. Although trained on point clouds rendered in PyBullet from the Block ID training split, the model operates directly on real point clouds captured before each regrasp. Selected placement proposals are executed on the robot without a complete object mesh or prior validation in simulation.

For each regrasp, the placement prior generates five proposals from the current partial observation. Each predicted normal is converted into a target object orientation using Rodrigues' formula to compute the minimum rotation that aligns it with the gravity direction $\mathbf{g}$. We assess graspability with GraspGen-X~\cite{graspgenx2026} and motion feasibility with cuRoboV2~\cite{curobo_v2}. Because density ranking does not account for information gain, we manually select among the feasible proposals to increase the visual information accumulated across reorientations. In particular, we exclude the initial placement at the second step to avoid revisiting it. After the first reorientation, we capture a new point cloud and repeat the process from the updated observation to execute the second reorientation. We register and merge the point clouds from the initial, intermediate, and final poses into a cumulative point cloud. The two reorientations expose surfaces unobserved in earlier views, illustrating the iterative accumulation of visual information through sequential exploratory regrasping.

\section{Conclusion} \label{sec:conclusion}
We presented FlipToSee, a probabilistic framework for predicting alternative stable placements from a single-view point cloud by modeling support normals with a vMF-MDN on $S^2$. Deterministic mode extraction forms a compact candidate set, and robustness-aware reranking prioritizes placements robust to angular perturbations. Simulation experiments show that the vMF-MDN placement prior improves placement reliability and mode coverage over its GMM-MDN counterpart, while the robotic demonstration shows that the learned placement prior can be integrated with grasp and motion planning for sequential exploratory regrasping. FlipToSee focuses on placement stability rather than information gain; future work will incorporate visibility objectives to select placements that maximize information gain during active visual exploration.


\bibliography{ref}

\end{document}